Evaluating Retrieval Augmented Generative Models for Document Queries in Transportation Safety

C.A. Melton, A. Sorokine, S. Peterson

Oak Ridge National Laboratory, Oak Ridge, TN, United States

National Security Sciences Directorate

ABSTRACT

Applications of generative Large Language Models (LLMs) are rapidly expanding across various domains, promising significant improvements in workflow efficiency and information retrieval. However, their implementation in specialized, high-stakes domains such as hazardous materials transportation is challenging due to accuracy and reliability concerns. This study evaluates the performance of three fine-tuned generative models—ChatGPT, Google's Vertex AI, and ORNL Retrieval-Augmented Generation augmented LLaMA 2 and LLaMA in retrieving regulatory information essential for hazardous material transportation compliance in the United States. Utilizing approximately 40 publicly available federal and state regulatory documents, we developed 100 realistic queries relevant to route planning and permitting requirements. Responses were qualitatively rated based on accuracy, detail, and relevance, complemented by quantitative assessments of semantic similarity between model outputs. Results demonstrated that the RAG-augmented LLaMA models significantly outperformed Vertex AI and ChatGPT, providing more detailed and generally accurate information, despite occasional inconsistencies. This research introduces the first known application of RAG in transportation safety, emphasizing the need for domain-specific fine-tuning and rigorous evaluation methodologies to ensure reliability and minimize the risk of inaccuracies in high-stakes environments.

INTRODUCTION

Applications of text generative Large Language Models (LLM) are being employed on an ever-widening array of domains. Their ability to ingest large corpora and provide queried response domain specific text (e.g., user manuals, regulatory documents) provides a user the opportunity to potentially expedite workflows that enhance productivity, save time and ultimately funding. However, as groundbreaking and shiny as these technologies might appear, generative LLM's are not all created equal. Although their performance can seem impressive at times, some generated responses may be filled with false information, hallucinations, or miss the mark with the depth of responses. These inconsistencies create significant roadblocks in ubiquitous adaption, especially when incorrect information can lead to safety or security issues.

One domain of particular interest is the transportation of hazardous materials (HM). HM transport contractors in the United States must adhere to hundreds of Federal and State regulations that specify practices including route planning and specific permits that must be held by the transporters. For example, many states have restricted and preferred routes depending on the category of material. This fact is especially important because many U.S. States have designated agent who deals with permit requests. As one could image, the understanding and knowledge of such regulatory documents, as well as information retrieval can be cumbersome and time consuming for a human.

Our study focused on information retrieval from regulatory requirements for HM transportation in the United States, where contractors must comply with numerous Federal and State regulations, including route planning and permits. To address the challenges of understanding and retrieving regulatory information, we compiled approximately 40 publicly available regulatory documents and fine-tuned three different generative models including a fine-tuned ChatGPT, fine-tuned Vertex AI, and an ORNL RAG-augmented LLaMA 2, and LLaMA3.

This work is novel and impactful due to the scarcity of subject matter experts in the transportation safety domain. It is not feasible to rely on standard AI models like a native ChatGPT or Gemini, as the field requires precise metrics and a clear chain of custody for responses. This fact holds especially true considering that inaccuracies could have severe safety and/or financial repercussions. Furthermore, typical LLM or RAG evaluation methods fall short because they often rely on evaluating materials similar to Wikipedia, which cannot ensure the precise and authoritative responses needed for this domain. To our knowledge, this is the first application of Retrieval-Augmented Generation (RAG) within transportation safety, providing a solution that meets the stringent demands for accuracy, reliability, and traceability.

The paper is structured as follows: We provide the reader with necessary background information regarding the evolution of LLMs, and other research relevant to this study. We then provide a thorough describe of methodologies and data used to investigate these phenomena as well as provide our results. Lastly, we will discuss potential implications of our results and conclude with a brief overview of the work conducted.

BACKGROUND

The development of LLMs has seen significant advancements since the introduction of BERT (Bidirectional Encoder Representations from Transformers) in 2018, which marked a turning point in natural language processing (NLP). BERT (Devlin et al., 2019), developed by Google, leveraged a transformer architecture that allowed for bidirectional comprehension, thus providing a more nuanced understanding of context compared to prior unidirectional models. Its key innovations (e.g. masked language modeling and next sentence prediction) enabled the BERT to excel in a wide range of NLP tasks. Following BERT, numerous variations of encoders and encoder-decoders based LLMs were developed in relatively quick succession, each improving upon BERT's foundation by optimizing training techniques or reducing model size while maintaining performance or expanding to multi-billion parameter models. These models have significantly advanced applications such as sentiment analysis, topic modeling, text classifications and summarization, and human speech-emulative generative text interactions. More recently, the NLP field has shifted toward even larger and more generalized models, such as GPT- 4, LLaMA, Gemini, and many others which utilize the transformer architecture at an unprecedented scale (Radford et al., 2019). These models are trained on vast datasets and demonstrate remarkable versatility across diverse tasks with minimal fine-tuning. As LLMs continue to evolve, their increasing complexity and capability reflect the broader shift towards models that can perform a wide range of tasks with greater accuracy and contextual understanding. This trajectory illustrates how the field has rapidly transitioned from task-specific NLP models to highly flexible, generalized language models, thereby expanding their applicability across different domains, including healthcare, social media analysis, and public policy development

Nowadays, several options are available for users to utilize generative pre-trained transformers models (GPT) that have been trained on vast corpora of text (i.e., the Internet). Free access is available to users limited to lower-tiered performance and less flexibility for fine-tuning with custom data (e.g., Google's Bard/Gemini, OpenAI's ChatGPT, Meta's LLaMA). Subscription-based services cater to various industries and give the user capabilities to fine-tune a GPT model with their own custom data set. For example, a user can upload a PDF, text file or image, and query the GPT model for the information they seek. In technical terms, several approaches towards incorporating users' data into LLMs. One of the common approaches is fine-tuning in which case model weights are modified by retraining on a small set of user-specific documents (Howard and Ruder, 2018). The second is RAG, a combination of document retrieval with a generative model to create the final answer (Lewis et al., 2020). Some models provide users the ability to preload a collection of the documents into the model context. Here we investigate and compare efficiency of each of these approaches.

METHODOLOGY

*Description of Data*

We gathered 40 documents related to transportation safety regulations in the United States to use in our RAG vector database, as well as ChatGPT customization, and Vertex AI's Search and Conversation. These documents consisted of state and federal regulations on information including but not limited to restricted and preferred travel routes, hazardous materials classifications, transportation security, permitting requirements, and authority contact information (see Table 1 and Table 2). The documents come from two primary sources, the International Atomic Energy Agency (IAEA) and the United States federal government. The federal regulatory documents used in the study are found in the Code of Federal Regulations (CFR) covering regulations and recommendations from the Nuclear Regulatory Commission (NRC) and the Department of Energy (DOE), found primarily in 10§CFR, and the Department of Transportation (DOT) in 49§CFR. We then developed 100 queries from the perspective of a route planner and evaluated the models' responses based on five specific criteria.

*Table 1. Document statistics for fine-tuning and RAG*

| **Statistic** | **Value** |
|---|---|
| Total words | 1,026,515 |
| Average word length | 5.14 characters |
| Total characters | 5,277,482 |

*Table 2. Most frequently used words (excluding stop words and numbers)*

| **Word** | **Frequency** |
|---|---|
| security | 8,858 |
| materials | 7,155 |
| nuclear | 6,287 |
| radioactive | 5,731 |
| transport | 4,927 |
| package | 3,444 |
| safety | 3,066 |
| requirements | 299 |

*Description of Vertex AI by Google*

Vertex AI by Google Cloud is a suite of machine learning tools designed to facilitate the development and deployment of artificial intelligence models. Among the features are significant capabilities in NLP. Built on top of Google's transformer architectures, such as BERT and its variants, Vertex AI's NLP tools are designed to aid in information retrieval, and text generative response, popular with customer service chat bots. These models utilize deep neural networks and attention mechanisms to interpret and process complex language structures, enabling tasks such as translation, summarization, and question answering. Over successive iteration, Vertex AI has incorporated fine-tuning mechanisms and hyperparameter optimization, allowing for the integration of domain-specific knowledge. Essentially, users can upload a series of documents to a Google Cloud bucket that acts as a database. The platform's design supports large-scale language processing, making it suitable for diverse industries from unstructured data. This functionality provides interpretability features through model explainability tools, ensuring transparency in predictions and references user documents when providing results (Google Cloud, 2023).

*ChatGPT*

ChatGPT by Open AI is now a series of generative language models, built on the GPT (Generative Pre-trained Transformer) architecture. Starting with GPT-1 and progressing through GPT-4o1, each iteration has significantly increased in size and capability, with GPT-4o1 featuring extensive parameter scaling to enhance text generation and contextual understanding. These models utilize deep neural networks and self-attention mechanisms to capture complex language patterns, enabling them to perform a wide range of tasks such as translation, summarization, and question-answering. Successive versions, including GPT-3.5 and GPT-4, have introduced improvements in fine-tuning, alignment with human feedback, and multi-modal processing, allowing the models to handle both textual and visual inputs effectively. ChatGPT's scalable and flexible design has made it a leading tool in NLP, widely applied in industries such as customer service, content creation, and education, while also offering interpretability through feature attribution techniques (Open AI, 2024).

*LLaMA 2 and LLaMA 3*

Meta's LLaMA (LLMl Meta AI) series, particularly LLaMA 2 and LLaMA 3, uses advancements in scalable transformer-based architectures. LLaMA 2 introduced models ranging from 7 billion to 70 billion parameters, trained on large and diverse datasets with optimizations such as mixed-precision training and gradient checkpointing to enhance computational efficiency. Building on these parameters, LLaMA 3 incorporates multi-modal capabilities by integrating convolutional neural networks (CNNs) for vision processing, enabling the handling of both textual and visual data. Furthermore, LLaMA also leverages advanced reinforcement learning techniques, including reinforcement learning from human feedback, to adapt more effectively to changing contexts without extensive retraining. Additionally, LLaMA 3 employs sparse attention mechanisms and low-rank factorization to reduce memory usage while maintaining high performance, making it well-suited for complex applications in fields such as information retrieval, healthcare, finance, and automated content generation (Meta, 2024).

*Retrieval Augmented Generation*

RAG is an NLP framework that uses the strengths of retrieval-based methods with generative models to enhance the quality and accuracy of generated content. By integrating a retrieval component, RAG uses external databases to fetch information sought by a user in near real-time. The generative model then has the capability to produce more informed and contextually appropriate responses. This aspect is especially important to users hoping to seek newer information or information not included in a models original training. RAG employs sophisticated techniques such as vector embeddings and attention mechanisms to ensure that the retrieved data seamlessly complements the generative process, thereby assisting to mitigate issues such as hallucination — enhancing the reliability of the output. Depending on the quality of the input data, RAG models can achieve better performance in tasks that require both extensive domain knowledge and access to large amounts of information, making them highly suitable for applications in domains such as healthcare, finance, customer support, and transportation safety (Lewis et al., 2020).

*Architecture of RAG Implementation*

In this study, we implemented the RAG architecture by combining several open-source software packages and language models. For the retrieval component, we utilized the llama index software and the BGE Embedding model, as described in (Liu, 2022). The generative component was implemented using the ollama software package, specifically employing the Llama2 and Llama3 models with 7 billion parameters. This configuration provided a robust framework for our research, allowing us to explore the capabilities of the RAG architecture using off-the-shelf hardware without using cloud-based services.

*Evaluation Methodology*

In the evaluation of LLMs, conventional methodologies predominantly focus on general questions unrelated to specific knowledge domains. However, developing a domain-specific evaluation technique poses a formidable challenge. The model is expected to provide answers on topics beyond its training data, specific to a document collection, and compliant with intricate rules and regulations. Moreover, these answers must be accurate and precise, as incorrect responses may have legal, financial, or environmental consequences. Existing evaluation methods primarily rely on curated question sets with expected answers, evaluation by other models, question generation from templates and other models, and human feedback. In compliance-regulated environments such as healthcare, law, or hazardous material transportation, the subject of our study, the current state of the art is limited by the need for evaluation by Subject Matter Experts (SMEs). As observed in previous research, SMEs are a scarce and costly resource, making their recruitment challenging. To address this issue, we have developed an evaluation methodology that minimizes SME effort while enabling the evaluation of multiple models, updates to document collections, and variations in question sets.

We developed a list of 100 questions related to information that could be found within the collected documents gathered for this study. To ensure a more authentic series of queries, these queries were developed with the assistance of an expert in the transportation safety domain.

To qualitatively evaluate generated responses, our team developed a rating system based on five criteria to test the performance of four models (ChatGPT 4.0, Gemini, and LLaMA 2 and 3 plus RAG). Each answer was rated on a scale of 1 to 5, with 1 being the worst answer and 5 the best. The numerical ratings were assigned according to the following criteria:

> 5) Delivers the queried information accurately and directly without hallucination, providing a high level of detail.

> 4) Delivers the queried information accurately, but the information may be obscured by superfluous details or filler text, or it could benefit from more thorough detail.

> 3) Provides a broad, high-level response to the query, often missing the specific point but still offering some valuable information.

> 2) Provides a broad, high-level response to the query, similar to a level 3 rating, but fails to deliver any useful information.

> 1) Fails to provide any relevant information regarding the intended query, or the model experiences complete failure, including hallucinations.

Query responses were then blindly evaluated based on these defined metrics by the three authors of this work. Scores were then averaged after evaluations were completed (see Fig. 1).

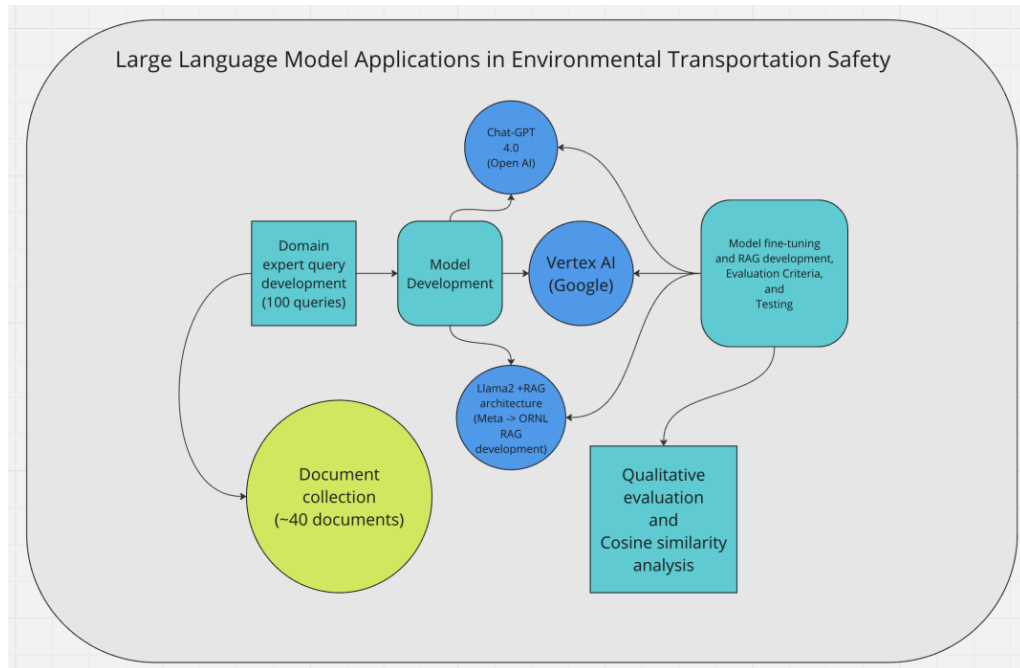

Figure 1. Flowchart describing methods used in this study.

Finally, we calculated the semantic similarity between answers to the same questions generated by different models. To achieve this, we employed the same embedding model utilized in the RAG retrieval step and applied the cosine similarity measure to compare embedding vectors. This approach enabled us to obtain objective similarity measures between the models and variations in the document collections. Furthermore, it helped reduce the evaluation burden on subject matter experts (SMEs) by eliminating the need to reevaluate semantically similar answers.

RESULTS

*Qualitative Analysis*

Among the models we tested, the ORNL LLaMA 2 plus RAG scored the highest, with an average score of 4.03, and stood out as the top performer. It generally delivered well-constructed answers, demonstrating a good understanding of the queries. However, despite its high average score, LLaMA 2 + RAG had a few instances where it provided contradictory information. This occurrence raises concerns about consistency in its output, particularly in scenarios that require a high level of reliability.

Google's Vertex AI scored second best, which scored an average of 3.41. Vertex AI demonstrated the ability to offer a decent level of detail in its responses, which could be useful in a variety of contexts, particularly when depth and explanation are necessary. However, its performance was not without fault. Similar to LLaMA 2 + RAG, Vertex AI also occasionally provided false or misleading information. This shortcoming diminished its overall effectiveness, though it still ranked higher than ChatGPT-4 due to its relatively more detailed responses.

ChatGPT-4, on the other hand, had the lowest average score at 3.03. Its responses were often described as "high-level," meaning they were more general or abstract than detailed or specific. These responses could represent potentially dangerous failures in situations that require precise answers. Furthermore, ChatGPT-4 exhibited multiple instances of false information, which affected its credibility and reliability in comparison to the other models. While still functional, it performed the weakest of the three models (see Appendix Table A)

*Similarity Estimation Between the Models*

Average semantic similarity between the answers to the same questions produced by different models is shown in Figure 3. The most similar answers were produced by the RAG architecture employing llama2 and llama3 models, the most dissimilar answers were created by llama3 and ChatGPT 4. However, the differences in average similarities are not high. Fig. 2. shows distribution of similarities between the pairs of models. All pairs of models show a large number of similar and very similar answers. Number of answers with large differences stays below 10% with the exception of the Chat GPT4 vs. Vertex AI pair in which case about a quarter of all answers shows high semantic dissimilarity. Examples of similar and dissimilar answers according to the used method is shown in Table A in Appendix.

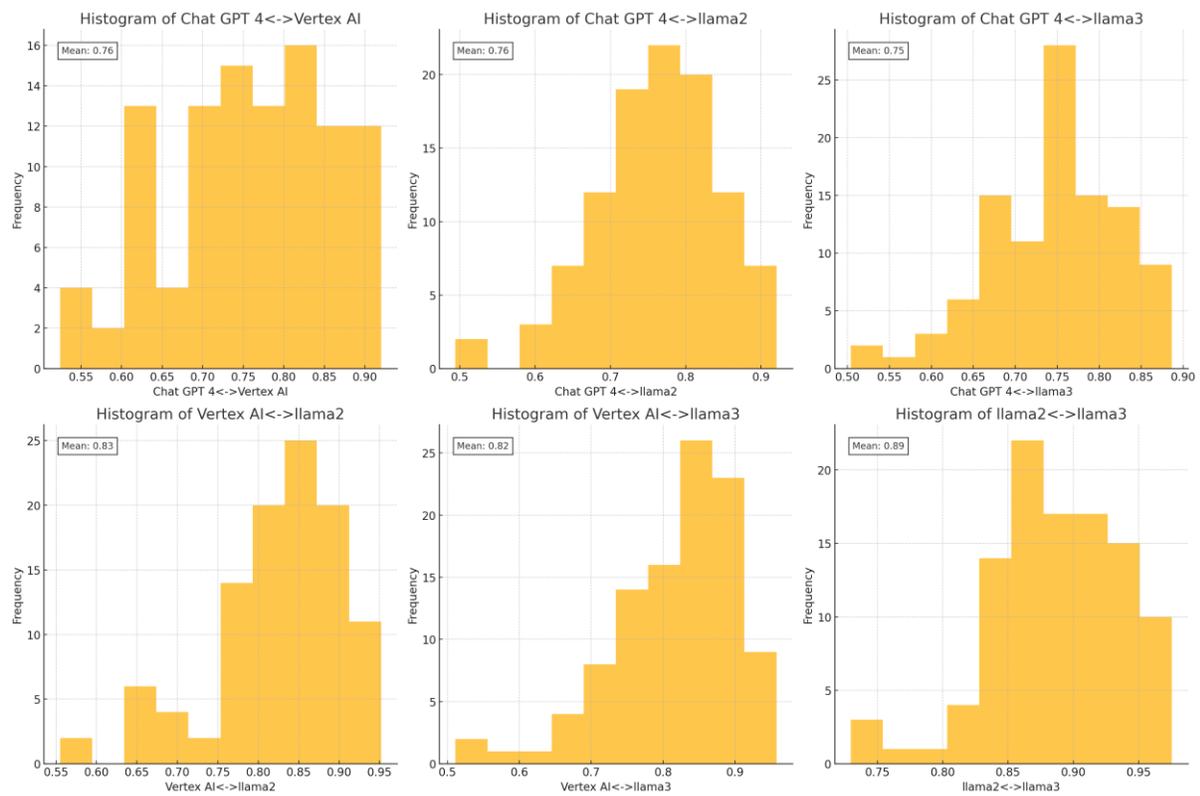

*Figure 2. Distributions of similarity between the pairs of models*

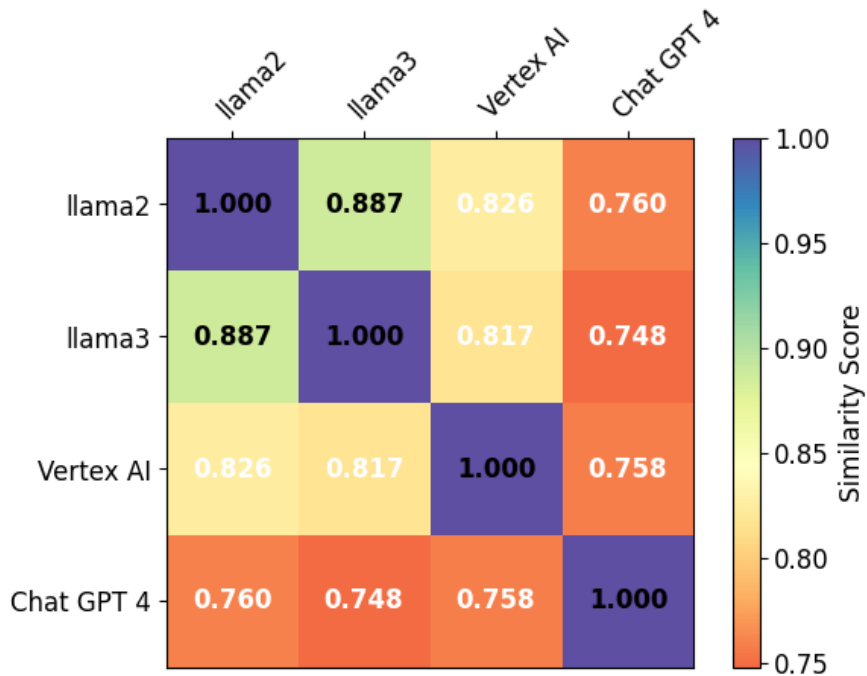

*Figure 3. Model similarity ranking matrix based on average*

DISCUSSION

*Qualitative Interpretation*

Overall, the evaluation suggests that while all three models have their strengths, none are perfect, with issues ranging from occasional contradictions to false information. LLaMA 2-3 plus RAG shows the most promise with its generally strong performance, but the occasional inconsistencies highlight an area for potential improvement. For example, when LLaMA asked this question:

"Are there special regulations under Title 49 for highway or truck shipments of radioactive materials?", LLaMA responded:

"Based on my analysis of the provided documents, there are no specific regulations under Title 49 for highway or truck shipments of radioactive materials. The regulations for transporting radioactive materials in the United States are primarily governed by the U.S. Department of Energy (DOE) and the U.S. Nuclear Regulatory Commission (NRC). These regulations can be found in CFR Titles 49 and 10, respectively. However, it's worth noting that there may be additional requirements or guidelines for transporting radioactive materials on highways or trucks, depending on the specific type of material being transported and the mode of transportation. It's important to consult with relevant authorities and experts in this field to ensure compliance with all applicable regulations and safety standards." Which unfortunately is false. ChatGPT and Vertex AI provided more accurate responses to this query.

Vertex AI offers more detail but could benefit from better accuracy. While ChatGPT 4o, though appearing capable, lags behind in both precision and depth, requiring refinement to compete more effectively with the other models.

*Semantic Similarity*

Quantitative assessment of semantic similarity between responses to identical questions generated by various models or derived from distinct document collections has revealed the sensitivity of this method in detecting even subtle variations in responses. Within the scope of this study, instances of semantically dissimilar responses rarely surpassed 10% of the total question count. Consequently, the workload associated with SME reevaluation of responses can be considerably diminished in comparison to reassessing the entirety of responses.

*Evaluation Needs*

This work emphasizes the importance of thoroughly evaluating responses generated by LLMs, regardless of the model chosen. While significant advances have been made since the introduction of BERT (Devlin et al., 2018), certain industries should exercise extreme caution when relying on LLMs. In sectors such as environmental transportation safety, or healthcare, errors can be a matter of life and death, or at the very least, result in significant regulatory or financial repercussions. For industries where accuracy is crucial, domain-specific LLMs fine-tuned on curated datasets provide a more reliable solution. The persistence of hallucinations across many models should also motivate NLP engineers to carefully consider the quality of data used to train base models. Hallucinations, where models generate incorrect or fabricated information, remain a critical issue, particularly in open-ended tasks where the model attempts to fill in gaps with plausible but erroneous data. Training a model on internet sources or "all of Wikipedia" exposes it to a considerable amount of misinformation. While Wikipedia provides a wealth of factual content, its susceptibility to edits by anyone introduces a degree of unreliability. Despite improvements in Wikipedia's quality (Steinsson, 2024), many other sources—such as news articles, media content, and social media—are riddled with misinformation, disinformation, and biases. For models operating in high-stakes environments such as healthcare or finance, curating high-quality datasets from peer-reviewed journals, regulatory documents, and verified repositories is paramount.

To combat this problem, approaches similar to RAG or the integration of external knowledge bases can aid in obtaining factual information, where users can at least tune hyperparameters and obtain metrics, and references to the documents used in the RAG database. And even though the models we tested in our study had some successes, none would be recommended for full implementation in the domain of environmental transportation safety at the current state of the art.

CONCLUSIONS
The use of generative LLMS continue to be explored, offering the potential to streamline workflows, enhance productivity, and save both time and resources. These models, capable of processing extensive corpora and generating domain-specific text, hold promise for fields requiring detailed and accurate information retrieval, such as the transportation of HM. However, despite their potential, generative LLMs often produce inconsistent results, including false information and hallucinations, which can hinder their widespread adoption, particularly in safety-critical areas. Our study focused on the regulatory requirements for HM transportation in the United States, where contractors must comply with numerous Federal and State regulations, including route planning and permits. We compiled approximately 50 regulatory documents and fine-tuned three different generative models with this data and developed 100 queries from the perspective of a route planner and evaluated the models' responses based on five specific criteria. Though no model was perfect in our study, the results showed that RAG augmented LLM performed best when compared Vertex AI and ChatGPT 4o. Our findings highlight both the potential and limitations of using generative LLMs for regulatory compliance in HM transportation. Future work efforts

APPENDIX

*Table A. Example of similar and dissimilar answers*

| Question | Model 1 Answer | Model 2 Answer |
|---|---|---|
| *What are the relevant regulations for transporting hazardous materials?*<br><br>*Cosine similarity of the answers: 0.975* | Llama 2<br><br>The relevant regulations for transporting hazardous materials, as mentioned in the provided documents, are:<br><br>1. IAEA Regulations for the Safe Transport of Radioactive Material (IAEA Safety Standards Series No. SSR-6, 2012 Edition) [1]<br>2. NRC Course H-308: Safeguards and Security [4]<br>3. 49 CFR 172 Subpart I [3]<br><br>The documents mention that these regulations apply to the transport of radioactive material by all modes on land or water, or in the air, including transport that is incidental to the use of the radioactive material. The regulations establish standards of safety that provide an acceptable level of control of the radiation, criticality and thermal hazards to persons, property and the environment associated with the transport of radioactive material.<br><br>The IAEA Regulations [1] provide guidance on how to comply with and demonstrate compliance with the requirements of the regulations. The NRC Course H-308 [4] provides additional guidance on security aspects of hazardous materials transportation, while 49 CFR 172 Subpart I [3] outlines major aspects of security plans and training for | Llama 3<br><br>Based on the provided documents, here are the relevant regulations for transporting hazardous materials:<br><br>1. IAEA Regulations for the Safe Transport of Radioactive Material (the Transport Regulations) [1] - These regulations apply to the transport of radioactive material by all modes on land or water, or in the air, including transport that is incidental to the use of the radioactive material.<br>2. 49 CFR 172 Subpart I - This regulation provides requirements for transporting hazardous materials, including security plans and training.<br><br>Additionally, the following documents provide guidance on compliance with these regulations:<br><br>1. Advisory Material for the IAEA Regulations for the Safe Transport of Radioactive Material [2] - This document provides guidance to users on how to comply with the requirements of the Transport Regulations.<br>2. NRC Course H-308: SAFEGUARDS and SECURITY - This course covers the transportation of radioactive material, including security plans and training.<br><br>Please note that these regulations and guidelines are specific to the transport of radioactive materials and may not be |
| | transportation of radioactive material.<br><br>Please note that these regulations may be subject to change, and it is important to consult the most up-to-date versions of these regulations before engaging in any transportation activities. | applicable to other types of hazardous materials. |
| *When should I use a Radioactive Yellow III label?*<br><br>*Cosine similarity of the answers: 0.494* | When the transport index exceeds 10, and the surface radiation level exceeds 50 mrem/hr but does not exceed 200 mrem/hr. | The user asked: When should I use a Radioactive Yellow III label?<br><br>Based on the provided documents, a Radioactive Yellow III label is used for shipments of radioactive materials that have a total radioactive content activity level of 5000 or more TBq (Terrabecquerels). This level is specified in paragraph 540(c) of the Regulations.<br><br>Therefore, the answer to the user question is: You should use a Radioactive Yellow III label when the total radioactive content activity level of the shipment is 5000 or more TBq. |